\documentclass[10pt,twocolumn,letterpaper]{article}

\usepackage[pagenumbers]{wacv} 

\usepackage{float}
\usepackage{graphicx}
\usepackage{tabularx}
\usepackage{multirow}
\usepackage{booktabs}
\usepackage{subcaption}

\usepackage{url}
\usepackage{times}
\usepackage{xspace}
\usepackage{pifont}
\usepackage{enumitem}
\usepackage{microtype}
\usepackage{soul}

\usepackage{colortbl}
\usepackage[most]{tcolorbox}
\usepackage[dvipsnames]{xcolor}

\usepackage{tikz}
\usepackage{pgfplots}
\pgfplotsset{compat=1.18}
\usepgfplotslibrary{groupplots}
\usetikzlibrary{positioning,fit,calc}
\usetikzlibrary{arrows.meta,backgrounds,shapes.geometric}

\usepackage{bm,bbm}
\usepackage{mathtools}
\usepackage{algorithm,algorithmic}
\usepackage{amsmath,amsfonts,amsthm,amssymb}

\renewcommand{\paragraph}[1]{\vspace{.5em}\noindent\textbf{#1}}

\newcommand{\tts}{test-time scaling\xspace}

\newcommand{\mmdl}{multimodal diffusion language model\xspace}

\definecolor{pattsDark}{HTML}{1F2933}
\definecolor{pattsMid}{HTML}{6B7280}
\definecolor{pattsLight}{HTML}{E5E7EB}
\definecolor{pattsBlue}{HTML}{4B5563}
\definecolor{pattsRank}{HTML}{374151}

\definecolor{StepBlue}{RGB}{0,114,178}
\definecolor{StepOrange}{RGB}{230,159,0}
\definecolor{StepGreen}{RGB}{0,158,115}
\definecolor{StepMagenta}{RGB}{204,121,167}

\definecolor{LightGreen}{HTML}{E8F5E9}

\usepackage{pgfplots}

\pgfplotsset{compat=1.18}

\newcommand{\visok}{\textcolor{ForestGreen}{\ding{51}}}
\newcommand{\visno}{\textcolor{BrickRed}{\ding{55}}}
\newcommand{\visimg}[2]{\raisebox{-0.5\height}{\includegraphics[width=0.178\linewidth]{asset/appendix_visuals/#1_#2.jpg}}}
\newcommand{\visprompt}[2]{\parbox[b]{0.178\linewidth}{\centering\footnotesize\texttt{#1}\\[1pt]{\scriptsize\textit{#2}}}}
\newcommand{\vismethod}[1]{\parbox[c]{0.085\linewidth}{\raggedright\small\textbf{#1}}}
\newcommand{\visrow}[7]{%
  \vismethod{#1} & \visimg{00326}{#2} & \visimg{00236}{#2} & \visimg{00404}{#2} & \visimg{00414}{#2} & \visimg{00531}{#2} \\[-1pt]
  & #3 & #4 & #5 & #6 & #7 \\[3pt]}

\definecolor{wacvblue}{rgb}{0.21,0.49,0.74}
\usepackage[pagebackref,breaklinks,colorlinks,allcolors=wacvblue]{hyperref}

\def\wacvPaperID{462}
\def\confName{WACV}
\def\confYear{2027}

\title{PATTS: Parallel-Aware Test-Time Scaling for Thought-Image \\
       Generation in Multimodal Discrete Diffusion LLMs}

\author{
Yasong Dai\textsuperscript{\rm 1, 2},$\;$
Zeeshan Hayder\textsuperscript{\rm 1},$\;$
David Ahmedt-Aristizabal\textsuperscript{\rm 2},$\;$
Hongdong Li\textsuperscript{\rm 1, 3}
\\[0.2em]
\textsuperscript{\rm 1}Australian National University,$\;$
\textsuperscript{\rm 2}CSIRO Data61,$\;$
\textsuperscript{\rm 3}Amazon
}

\begin{document}

\newcommand{\stepblue}[1]{\textcolor{StepBlue}{#1 }}
\newcommand{\steporange}[1]{\textcolor{StepOrange}{#1 }}
\newcommand{\stepgreen}[1]{\textcolor{StepGreen}{#1 }}
\newcommand{\stepmagenta}[1]{\textcolor{StepMagenta}{#1 }}

\twocolumn[{
\maketitle
\begin{center}
\captionsetup{type=figure}
\begin{minipage}{\linewidth}
    \begin{minipage}{0.70\linewidth} 
    \subcaptionbox{
    PATTS trajectory search with reliability-based verification.
    }{
    \includegraphics[width=\linewidth]{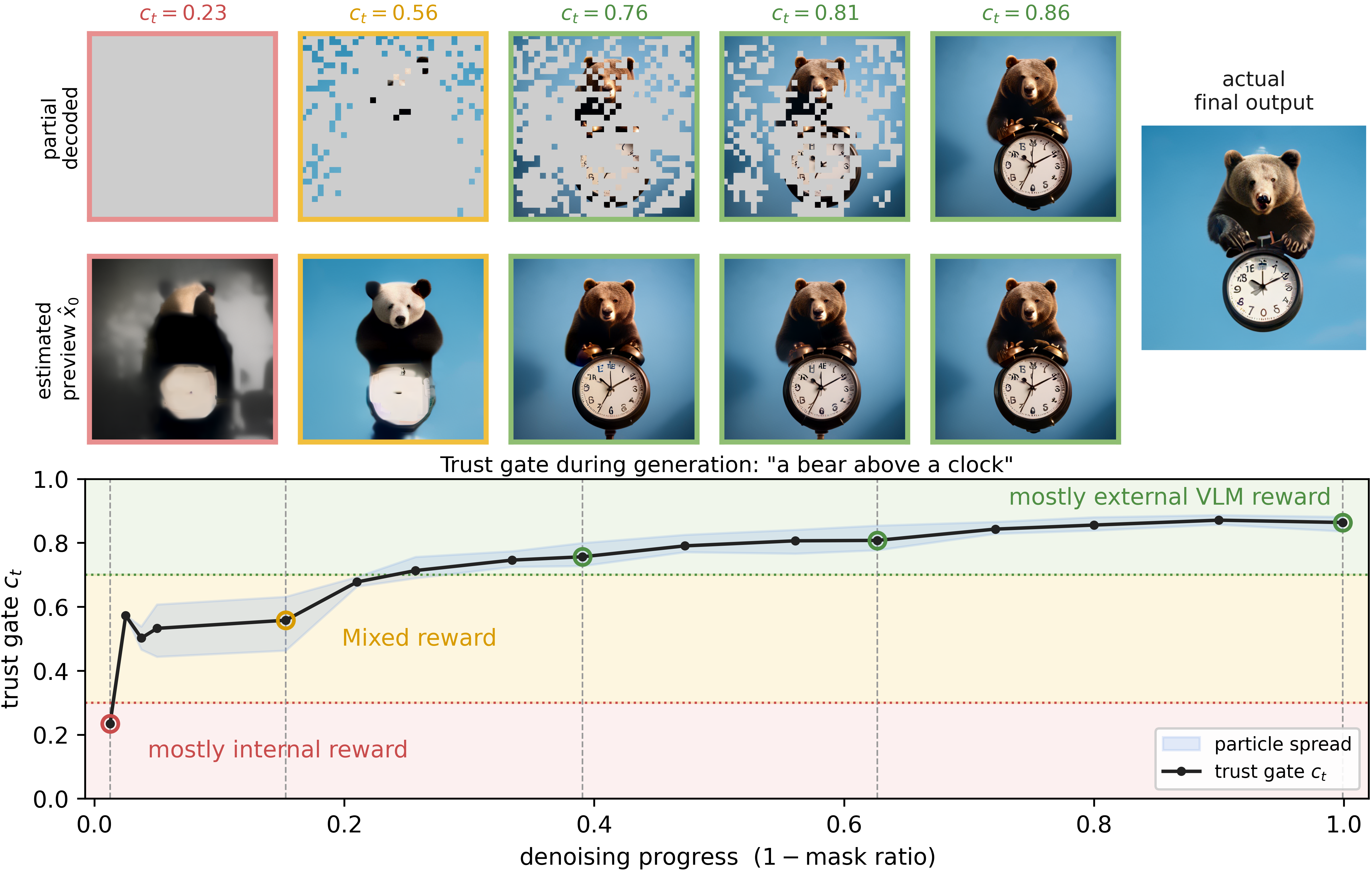}
    }
\end{minipage}
\hfill
\begin{minipage}{0.25\linewidth}
    \centering
    \includegraphics[width=0.75\linewidth]{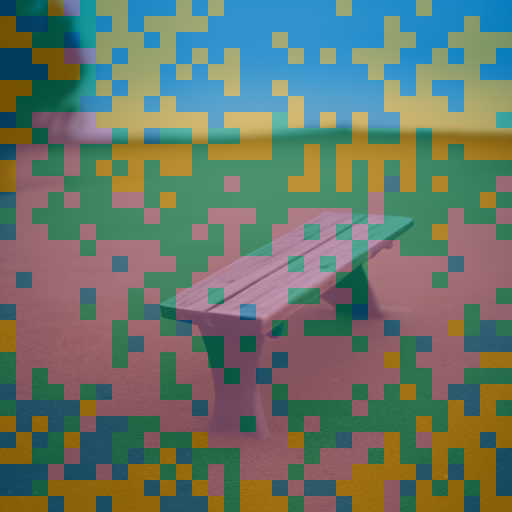}
    \caption*{\scriptsize
    \stepblue{\rule{1.2em}{0.8em} 0-3}\quad
    \steporange{\rule{1.2em}{0.8em} 4-7}\quad
    \stepgreen{\rule{1.2em}{0.8em} 8-11}\quad
    \stepmagenta{\rule{1.2em}{0.8em} 12-15} \\
    \stepblue{
    1. \textbf{Lighting}: The current image has shadows ...}
    \steporange{ light, which can enhance the mood and}
    \stepgreen{ create}
    \steporange{ a}
    \stepgreen{ more serene atmosphere}
    \steporange{. 2. \textbf{Color}}
    \stepgreen{\textbf{ Tone}}
    \steporange{\textbf{:} The colors in the question image are muted and somewhat}
    \stepgreen{ desaturated}
    \steporange{. The answer image}
    \stepgreen{ has a vibrant blue tint, which}
    \stepmagenta{ could be due to ...}
    \stepgreen{. The answer image}
    \stepmagenta{ has}
    \stepgreen{ a shallow}
    \stepmagenta{ depth}
    \stepgreen{ of}
    \stepmagenta{ field, focusing on}
    \stepgreen{ the bench ... 
    and the background}
    \stepblue{. 4. }
    \steporange{\textbf{and Texture}}
    \stepblue{\textbf{:} The}
    \steporange{ wood}
    \stepblue{ in}
    \steporange{ the}
    \stepblue{ answer image}
    \steporange{ looks slightly more worn and aged compared to the clean, new bench in the question image.}
    \stepblue{ ...
    By applying these adjustments, the image image will better reflect the transformation.} \\[3pt] {\footnotesize (b) Parallel thought-image generation.}
    }
\end{minipage}

\end{minipage}

\caption{{\bf Overview of PATTS.} 
A parallel diffusion multimodal LLM generates multiple partially decoded thought-image trajectories. At selected cut point, PATTS forms a predicted clean preview \(\hat z^{(t)}_0\) and combines model-internal reliability diagnostics with an external Vision-Language-Model (VLM) score. 
The state-dependent reliability weight $c_t(z_t)$ increases the influence of external semantic verification only when the intermediate preview is sufficiently stable to rank.
}
\label{fig:overview_PATTS}
\end{center}

\vspace{1em}
}]

\begin{abstract}
Test-time scaling (TTS) improves generative models by allocating additional inference-time computation to sampling, search, and verification.
Existing image generation approaches, however, typically rank completed images or image-centric intermediate states. We study TTS for parallel multimodal discrete diffusion language models, in which reasoning tokens and image tokens are decoded jointly within a single generation trajectory.
We first diagnose a limitation of conventional final-image selection: on GenEval, a CLIP-style image-prompt score is weakly aligned with within-prompt compositional correctness, leaving substantial headroom between verifier selection and oracle selection.
We then propose \textbf{PATTS}, a novel parallel-aware TTS framework that searches over intermediate thought-image states using cross-modal verifier signals. PATTS uses a state-dependent reliability weight, derived from model-internal trajectory diagnostics, to modulate the influence of external semantic verification during generation.
In the fixed-$N=32$ evaluation on MMaDA-Parallel, the complete PATTS configuration reaches a GenEval score of $0.844$, compared with $0.628$ without TTS, $0.787$ for Best-of-N, and $0.810$ for FKD.
These results show that reliability-aware trajectory verification can improve test-time selection for parallel thought-image co-generation under the evaluated search budget.
Our code release is at \url{https://github.com/Ringo-Star/PATTS.git}
\end{abstract}   
\section{Introduction}
\label{sec:intro}

Test-time scaling (TTS) improves generation by spending extra computation at inference time rather than retraining the model. In visual generation, this extra budget is usually allocated along image-centric axes: more denoising steps, more random trajectories, or best-of-$N$ selection with a final-image VLM verifier model~\cite{ma2025inference,he2025scaling,sundaresha2026verifier}. These strategies are effective when a completed image is the natural unit of search. However, they are less complete for multimodal diffusion language models, where generation can expose a richer trajectory than a sequence of final images.

Conventional CLIP-style prompt-image scores primarily capture global semantic alignment and may fail to distinguish images that satisfy fine-grained compositional constraints from those that do not. On GenEval, CLIP-based search improves the average correctness of a 16-trajectories TTS only from 86.51\% to 87.34\%, while an oracle selector reaches 96.75\%. This substantial gap motivates exploring additional information beyond final-image similarity for guiding test-time search.
Parallel thinking-generation models make this limitation explicit. In MMaDA-Parallel~\cite{tian2026mmadaparallel}, textual reasoning tokens and image tokens are decoded jointly inside a single masked diffusion process, as shown in Figure~\ref{fig:overview_PATTS}. This creates a new inference-time scaling opportunity: instead of verifying the final image, a search method can inspect partially decoded thought-image states and select trajectories using their evolving cross-modal information.

However, exploiting these intermediate states introduces a new challenge: their semantic content may be unreliable before denoising is sufficiently advanced. Early previews can be unstable, while late previews are easier to verify but leave less room to redirect generation. Moreover, the reasoning and image states may become misaligned, producing plausible reasoning traces alongside images that violate spatial relations or fine-grained attributes. The central question is therefore not simply how to sample more candidates, but when an intermediate thought-image state is reliable enough to guide search.

We study this problem in parallel multimodal discrete diffusion LLMs and propose \textbf{PATTS}, a parallel-aware test-time scaling framework for thought-image generation. PATTS maintains multiple denoising trajectories and forms predicted clean thought-image previews at selected cutpoints. Rather than applying external semantic verification uniformly across denoising steps, PATTS estimates the reliability of each intermediate state using lightweight model-internal diagnostics, include token confidence, uncertainty, preview stability, and cross-modal agreement. A state-dependent trust gate then adaptively combines internal diagnostics with external VLM verification, reducing reliance on unstable intermediate previews while emphasizing external verification only in late stage of generation process.

Although intuitive straightforward, our experiments demonstrate the effectiveness of reliability-aware trajectory search. On MMaDA-Parallel, enabling parallel reasoning without verification reduces GenEval score from 0.65 to 0.62, with the largest degradation on position and attribute. Under a fixed $N=32$ evaluation, PATTS achieves 0.844, compared with 0.628 without TTS, 0.787 for Best-of-N, and 0.810 for FKD~\cite{zhang2026classic}.
The fixed-setting component breakdown further shows that trajectory-level search, reward routing, and state-dependent trust estimation contribute progressively to the final result, avoiding comparisons across different budgets or configurations.

Our contributions are:
\begin{enumerate}
    \item We identify empirical limitations of conventional final-image selection TTS for parallel thought-image generation, including weak alignment between CLIP final-image scores and compositional correctness, and image semantic degradation under unverified parallel thinking.
    \item We propose PATTS, a trajectory-level, reliability-based search framework that uses model-internal reliability diagnostics to weight external VLM verification of partially decoded thought-image states.
    \item We demonstrate improved compositional generation quality and inference efficiency over comprehensive text-to-image generation benchmarks and TTS baselines, and ablate the individual contributions of reward routing and state-dependent trust estimation.
\end{enumerate}

\section{Related Work}
\label{sec:relatedwork}

\paragraph{Test-time scaling for visual generation.}
Test-time scaling allocates additional inference compute to improve a frozen generator through sampling, search, or verification. In language models, this idea appears as increased reasoning or search budget~\cite{snell2024scaling,muennighoff2025s1}; In diffusion image generation, recent methods allocate compute to noise search, evolutionary search, path pruning, or verifier-guided candidate selection~\cite{huang2025selfcalibration,ma2025inference,he2025scaling,sundaresha2026verifier,yu2025ttsnap}. These approaches largely treat the image trajectory as the object of search and score either final samples or visual denoising states. We instead study a different search object: the joint thought-image trajectory exposed by parallel multimodal diffusion LLMs.

\paragraph{Thinking-aware multimodal diffusion models.}
Thinking-aware image generation asks whether explicit reasoning can improve synthesis and editing. Several recent systems condition generation on a reasoning trace or follow a sequential think-then-generate paradigm~\cite{guo2025imagecot,deng2025bagel,ai2026llada2}. In such pipelines, an error in the thought can propagate into the image, and the image state cannot usually correct the reasoning state once the thought is fixed~\cite{yang2026mmada}.
Parallel multimodal diffusion models change this interface: MMaDA-Parallel~\cite{tian2026mmadaparallel} decodes reasoning and image tokens jointly, while unified discrete diffusion generators such as Lumina-DiMOO and Muddit~\cite{xin2025lumina,shi2025muddit} show how masked-token diffusion can support multimodal generation. PATTS uses this parallelism as the testbed for trajectory-level \tts rather than treating the model architecture itself as the contribution.

\paragraph{Verifier-guided multimodal generation.}
Recent multimodal TTS work extends language-model-style inference scaling to unified multimodal models. However, general vision-language models offer limited verifier reward quality on complex and compositional image generation benchmarks~\cite{huang2025t2icompbench}. 
CLIP-style prompt-image scores~\cite{radford2021clip} are convenient for best-of-$N$ selection, but compositional benchmarks often require object-level counting, attributes, relations, and binding rather than global image--text similarity.
Recent multimodal TTS and refinement systems use stronger VLMs or unified models to generate, verify, and revise outputs through multi-round reasoning~\cite{chen2026unit,zhang2025omniverifier,xin2026dmllm-tts,jiang2026ttsp}. These methods mainly verify completed outputs or sequential refinement states. 
Our focus is on when partially decoded thought-image states in a parallel diffusion trajectory are reliable enough to verify.
We find that final prompt-image verifiers can select visually plausible samples while still failing to rank compositional correctness reliably. This motivates treating verification as a trajectory-dependent reliability problem rather than a fixed final-output scoring step.

\paragraph{Intermediate-state verification.}
Intermediate verification is difficult because intermediate states are not always semantically meaningful. In continuous diffusion models, a verifier often sees a noisy latent or a predicted clean image, so prior work use estimate of final output from current state~\cite{ma2025inference,he2025scaling,sundaresha2026verifier} or train a noise-aware verifier model~\cite{yu2025ttsnap}.
Discrete diffusion multimodal LLMs~\cite{austin2023d3pm} provide a different interface: an intermediate state is partially unmasked, and a clean preview can expose both predicted reasoning tokens and predicted image tokens~\cite{nie2025llada,xin2025lumina,shi2025muddit}. This makes prompt-thought, prompt-image, and thought-image consistency measurable before the final sample, but only when the preview is stable enough.
Existing intermediate-state verifiers do not exploit this structure and remain insufficient for the partially decoded thought--image states that arise in parallel thinking-aware dMLLMs. We address this reliability problem directly through a state-dependent trust gate rather than applying a fixed verifier schedule.
\section{Method}
\label{sec:method}

\begin{figure*}
    \centering
    \includegraphics[width=\linewidth,height=0.3\linewidth]{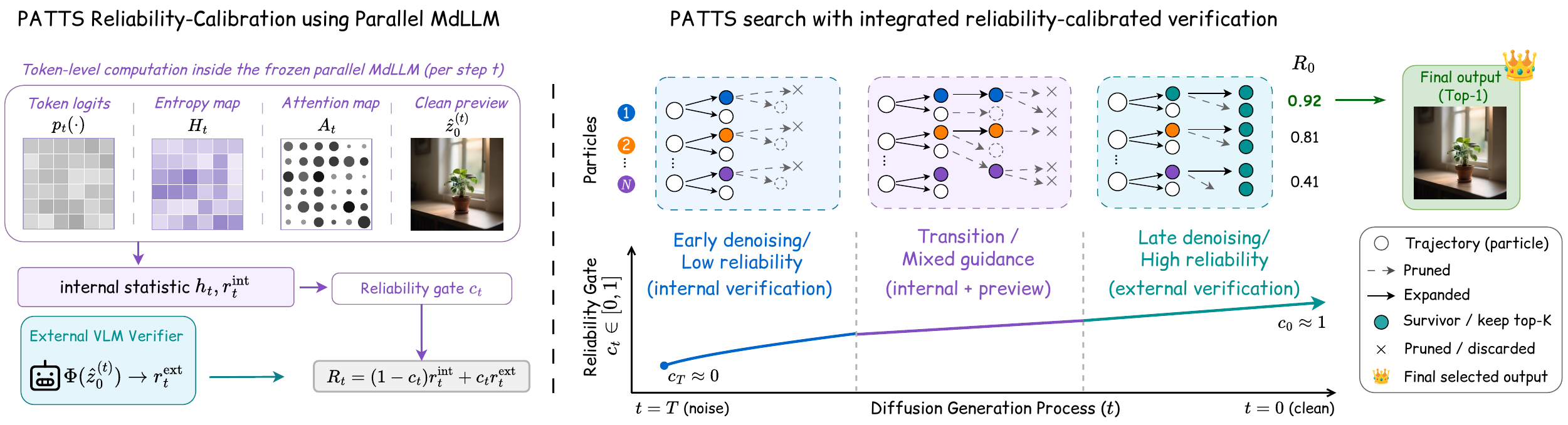}
    \vspace{-1em}
    \caption{{\bf Method Overview of PATTS.}
    \textbf{Left:} at each denoising cutpoint, the parallel MdLLM provides token logits, entropy statistics, attention patterns, and an estimated clean preview $\hat{z}^{(t)}_0=(\hat{y}^{(t)}_0,\hat{v}^{(t)}_0)$, which are converted into internal reliability features $h_t$. Then the reliability gate $c_t$ as well as $R_t$ will be calculated.
    \textbf{Right:} $c_t$ routes reward weight between internal trajectory reliability and an external VLM verifier. Based on reward ranking, trajectories are scored, pruned, and propagated. The final output is selected from the surviving trajectories.}
    \label{fig:method_main}
    \vspace{-0.5em}
\end{figure*}

\subsection{Preliminaries}
\label{subsec:preliminary}

\paragraph{Image-thought parallel multimodal diffusion generation.}
Following \cite{tian2026mmadaparallel}, in a multimodal Diffusion LLM, an intermediate state $z_t$ at denoising step $t$ is:
\begin{equation}
    z_t := (y_t,v_t \mid h_t),
    \label{eq:inter_states}
\end{equation}
where $y_t$ are reasoning tokens, $v_t$ are image tokens, $h_t$ refers to model internal information during intermediate timesteps. The final output is $z_0=(y_0,v_0)$. 
Text and image tokens are denoised in parallel but may follow different schedules, so partially decoded images can be viewed as checkpoints of a finer textual reasoning process.
For evaluation of intermediate image tokens $v_t$, a clean preview can be formed by filling still-masked positions with the $\text{argmax}$ logits. Decoding $\hat v_0^{(t)}$ then yields a paired thought-image estimate:
\begin{equation}
    \hat z_0^{(t)}=\hat z_0(z_t)=(\hat y_0^{(t)},\hat v_0^{(t)}),
    \label{eq:preview}
\end{equation}

\paragraph{Intermediate reward alignment.}
Let $p_{\theta}(z_0)$ be the clean-sample distribution of the model $\theta$ and $r(z_0)$ a final reward. A KL-regularized reward-aligned target is:
\begin{equation}
    p_{\text{target}}(z_0)=\frac{1}{Z}p_{\mathrm{\theta}}(z_0)
    \exp\!\left(\frac{r(z_0)}{\alpha}\right),
    \label{eq:target}
\end{equation}
with temperature $\alpha>0$. The ideal intermediate potential marginalizes over all clean samples:
\begin{equation}
    V_t(z_t)=\alpha\log\mathbb{E}_{p(z_0\mid z_t)}
    \left[\exp\!\left(\frac{r(z_0)}{\alpha}\right)\right],
    \label{eq:potential}
\end{equation}
which is intractable. Practical methods~\cite{he2024manifold,kim2025das} instead use a plug-in score on estimated clean view:
\begin{equation}
    \hat r_t(z_t):=r\!\left(\hat z_0^{(t)}\right).
    \label{eq:surrogate}
\end{equation}
However, this surrogate can be biased when previews are unstable, as we have already shown in Figure~\ref{fig:overview_PATTS}.

\subsection{PATTS overview}
\label{subsec:overview}

We propose PATTS, a test-time scaling framework for parallel-aware \mmdl{}s. Figure~\ref{fig:method_main} illustrates the PATTS pipeline. The key motivation of PATTS is that an intermediate thinking process is useful only when it can be faithfully estimated and remains aligned with the generated image. Otherwise, early reasoning can drift and hurt compositional accuracy. Therefore, PATTS treats verification as a \emph{reliability-weighted search problem}: PATTS first converts the partially masked state $z_t$ into a predicted clean thought--image preview $\hat{z}^{(t)}_0=(\hat{y}^{(t)}_0,\hat{v}^{(t)}_0)$. It then computes lightweight internal reliability diagnostics from the current trajectory and uses them to estimate a trust gate $c_t(z_t)$. This gate determines how much selection reward should come from internal  reliability versus an external VLM verifier applied to the clean preview.
\begin{equation}
    R_t^{\mathrm{PATTS}}(z_t)=
    \big(1-c_t(z_t)\big)\,\tilde r_t^{\mathrm{int}}(z_t)
    +c_t(z_t)\,\tilde r_t^{\mathrm{ext}}(z_t),
    \label{eq:patts_mix}
\end{equation}
where $c_t(z_t)\in[0,1]$, and $\tilde r_t^{\mathrm{int}},\tilde r_t^{\mathrm{ext}}$ are `standardized rewards. Trust gate values indicate whether the trajectory is confident, stable, and cross-modally coherent enough to be verified. High trust gate values shift verification reward weight to VLM semantic thought-image verification.

\subsection{Trust estimation from trajectory reliability}
\label{subsec:trust}

A standard intermediate reward sampler applies a pre-designed timestep schedule, which globally suppresses early rewards, but reliability in parallel multimodal diffusion also depends on the prompt, token uncertainty, preview stability, and text--image agreement. In addition, these methods do not give effective verification signals in early generation steps.

PATTS therefore estimates reliability from the current state: In test-time scaling setting, for $N$ trajectories $\{z_t^{(i)}\}_{i=1}^N$, trust is the probability that the intermediate ordering agrees with the final reward ordering:
\begin{equation}
    c_t=\Pr\!\left[
    \operatorname{sign}(\hat r_t^{(i)}-\hat r_t^{(j)})
    =\operatorname{sign}(r_0^{(i)}-r_0^{(j)})
    \right].
    \label{eq:trust_rank}
\end{equation}
This quantity cannot be computed online without completing many trajectories. PATTS therefore uses a state-dependent proxy ($c_t(z_t^{(i)})$), estimated from cheap trajectory features:
\begin{equation}
    \bm{h}_t^{(i)}=[H_t^{(i)},M_t^{(i)},\rho_t^{(i)},S_t^{(i)},A_t^{(i)},\mathrm{RankStab}_t^{(i)}].
    \label{eq:trust_feat}
\end{equation}
Here $H_t$ is normalized mean entropy over masked reasoning and image positions, $M_t$ is the mean top-two logit margin, and $\rho_t=|\mathcal M_t|/K$ is the remaining mask ratio. The clean preview defines the remaining features: Denoising stability is token overlap between adjacent previews, while cross-modal agreement is cosine similarity in the frozen shared embedding space, $A_t^{(i)}=\cos\!\big(\bar E(\hat y_0^{(t)}),\bar E(\hat v_0^{(t)})\big)\label{eq:agreement}$. Rank stability measures how much a particle's preliminary rank changes between adjacent cutpoints, where ranks are computed by $M_t^{(i)}-H_t^{(i)}+S_t^{(i)}$. The trust gate is
\begin{equation}
    c_t(z_t^{(i)})=\sigma\!\left(\omega_0+\omega^\top h_t^{(i)}\right),
    \label{eq:trust_gate}
\end{equation}
Here, $c_t^{(i)}$ is used as a state-dependent reliability weight for mixing internal and external scores, rather than as a calibrated probability of final ranking agreement. The feature normalisation and the hyperparameters $(\omega_0,\omega,\eta,\tau_{\mathrm{ext}},\beta)$ are fixed on a separate development set and kept fixed throughout evaluation, without prompt-specific tuning. 

\begin{algorithm}[t]
\caption{Trust-calibrated intermediate verification}
\label{alg}
\begin{algorithmic}[1]
\REQUIRE prompt $p$; generator $p_\theta$; verifier $\Phi$; particles $N$; cutpoints $T>t_1>\cdots>0$
\STATE initialize $N$ particles $\{z_T^{(i)}\}_{i=1}^N$
\FOR{each cutpoint $t$}
\FOR{each particle $i$}
\STATE predict clean preview $\hat z_0^{(i,t)}=(\hat y_0^{(i,t)},\hat v_0^{(i,t)})$
\STATE compute diagnostics $h_t^{(i)}$: entropy, margin, mask ratio, stability, agreement, rank stability
\STATE estimate trust $c_t^{(i)}=\sigma(\omega^\top h_t^{(i)})$
\STATE compute reward $r_t^{\mathrm{int}}(z_t^{(i)})$ and uncertainty $u_t^{(i)}$
\STATE if $c_t^{(i)}\!\ge\!\tau_{\mathrm{ext}}$: external reward $r_t^{\mathrm{ext}}(z_t^{(i)})$ from $\hat z_0^{(i,t)}$ via verifier logits (else reuse cache)
\ENDFOR
\STATE standardize $r_t^{\mathrm{int}},r_t^{\mathrm{ext}}$ across particles $\to \tilde r_t^{\mathrm{int}},\tilde r_t^{\mathrm{ext}}$
\STATE score $s_t^{(i)}=(1-c_t^{(i)})\tilde r_t^{\mathrm{int},(i)}+c_t^{(i)}\tilde r_t^{\mathrm{ext},(i)}-\beta u_t^{(i)}$
\STATE keep top-$K$ by $s_t$
\STATE update controller using rank stability and ESS
\STATE propagate surviving particles with frozen $p_\theta$
\ENDFOR
\STATE \textbf{return} final candidates $z_0$, optionally reranked by $\Phi$
\end{algorithmic}
\end{algorithm}

\subsection{Internal and external verification rewards}
\label{subsec:rewards}

\paragraph{Internal trajectory reward.}
When reliability is low, PATTS avoids applying semantic verification to unstable previews and instead scores trajectory reliability:
\begin{equation}
    r_t^{\mathrm{int}}(z_t) = \bm{\eta}^\intercal \bm{h}_t
    \label{eq:r_int}
\end{equation}
with coefficients $\bm{\eta}$. This reward does not assert semantic correctness and favors particles likely to become reliable enough for later semantic verification.

\paragraph{External semantic reward.}
When reliability is high, PATTS evaluates the clean preview with a thinking-aware verifier $\Phi$: The score is read as an affirmative probability from yes/no verifier logits,
\begin{equation}
    r^\text{ext}_t(z_t)=\frac{e^{\ell_{\mathrm{Yes}}}}{e^{\ell_{\mathrm{Yes}}}+e^{\ell_{\mathrm{No}}}}
    =\sigma(\ell_{\mathrm{Yes}}-\ell_{\mathrm{No}}),
    \label{eq:verifier_logit}
\end{equation}
which preserves cross-particle resolution better than decoded integer ratings. Because the verifier is expensive, it is evaluated only when $c_t\ge\tau_{\mathrm{ext}}$; otherwise the cached score is reused and its contribution remains small through Eq.~\eqref{eq:patts_mix}.

\paragraph{Standardized scoring.}
Internal and external rewards have different scales and variances. PATTS standardizes each reward across the current population before mixing: Thus $c_t$ controls relative influence rather than raw scale. If the external verifier returns nearly identical scores, $\sigma_t^{\mathrm{ext}}\to0$ and the standardized external term contributes little. Our default implementation keeps the top-$K$ particles by $R_t$ under a linearly decaying budget.

\subsection{Effective model-internal reliability signals}
\label{subsec:internal_reward_justification}

The internal reward in Eq.~\eqref{eq:r_int} is a reliability estimator, not a semantic verifier. Its signs follow from the masked-token posterior that produces the clean preview~\cite{lou2024sedd}. For each masked position $i$, let $p^i=p_\theta(\cdot\mid z_t)$, $p^i_{\max}$ be the preview-token probability, $H^i$ the predictive entropy, and $M^i=p^i_{(1)}-p^i_{(2)}$ the top-two margin. Since the masked-diffusion objective is cross-entropy, preview likelihood is controlled by $\sum_i\log p^i_{\max}$, and
\begin{equation}
    -\log p^i_{\max}\le H^i,
    \qquad
    -\log p^i_{\max}\le -\log M^i.
    \label{eq:entropy_margin_preview_nll}
\end{equation}
Low entropy and high margin therefore select previews that are high-likelihood under the frozen denoiser. The margin also has a robustness interpretation: the argmax token is unchanged under any $\ell_\infty$ probability perturbation smaller than $M^i/2$.

Mask ratio and denoising stability control plug-in reward bias. A high $\rho_t$ indicates many unresolved positions; high $S_t$ indicates that adjacent clean previews have converged. If the final verifier is locally smooth in token-space distance, stable previews are more likely to preserve final rankings, matching the trust criterion in Eq.~\eqref{eq:trust_rank}. Cross-modal agreement $A_t$ is specific to the parallel multimodal setting: bidirectional attention lets partial reasoning influence image tokens and partial image tokens influence reasoning tokens. Cosine agreement in the shared embedding space is a lightweight proxy for text--image dependence, in the same general form as contrastive mutual-information and CLIP-style alignment objectives~\cite{oord2018cpc,poole2019variational,radford2021clip}. Finally, $\mathrm{RankStab}_t$ directly measures whether particle orderings persist across cutpoints. Together, these signals route selection toward particles that are confident, stable, cross-modally aligned, and likely to become externally verifiable.

\section{Experiments}
\label{sec:experiments}

\subsection{Setup}
\label{sec:exp_setup}

\paragraph{Benchmark and metric.}
We use the GenEval compositional generation benchmark~\cite{ghosh2023geneval} with 553 prompts spanning six categories: single object, two object, counting, colors, position, and color attributes. Generated images are scored with the standard Mask2Former object detector~\cite{cheng2022mask2former}, and we report the GenEval overall score as the unweighted mean of the six category accuracies.
We additionally evaluate cross-benchmark robustness on T2I-CompBench~\cite{huang2025t2icompbench} and DPG-Bench~\cite{hu2024ella} using each benchmark's official evaluator and overall score.
Unless otherwise annotated, evaluations use 4 repeated runs to decrease evaluation instability.

\paragraph{Generators and configurations.}
We use the parallel thinking-generation model MMaDA-Parallel~\cite{tian2026mmadaparallel} as the primary generator.
GenEval scores for more diffusion generators~\cite{rombach2022ldm,shi2025muddit,yang2026mmada,xin2025lumina} are summarized in the supplementary material.
Unless stated otherwise, the reference configuration is thinking-aware decoding (32 image denoising steps, 128 text denoising steps, CFG 4.0, at resolution 512$\times$512).

\paragraph{Verifiers.}
We choose three verifiers in the analysis. For the thinking-aware generator we use a popular VLM model, Qwen2.5-VL-7B-Instruct~\cite{qwen2025qwen25}, which scores the prompt against the reasoning text, the final image, or both jointly.  A CLIP-style image-prompt score is the conventional choice for best-of-$N$ selection over completed images. A structured matching score (Oracle) that mirrors GenEval's object-level checks serves as a benchmark-aligned reference.

\begin{table*}[ht]
    \centering
    \small
    \caption{
    {\bf Quantitative performance comparison on GenEval with MMaDA-Parallel.} \textbf{NFE} counts forward passes through the image generator; \textbf{Ver.} counts {verified candidate previews} per prompt. \textbf{Latency} is end-to-end wall-clock latency per generated image. We report error bars only for the overall score. {\bf Best results} and \underline{second-best results} are annotated for each column. $\dagger$ reports re-implementation from original baseline on MMaDA-Parallel.
    }
    \label{tab:main_tts_comparison}
    \vspace{-0.2cm}
    \begin{tabular}{lllll|cccccc}
    
        \toprule
        
        \textbf{Method} & \textbf{NFE} & \textbf{Ver.} & \textbf{Latency} & \textbf{GenEval $\uparrow$} & Single & Two & Counting & Colors & Position & Attributes  \\
        
        \midrule

        No TTS
        & \textbf{161} & \textbf{0} & \textbf{9.9} & {0.628$_{\pm.037}$}
        & 0.975 & 0.657 & 0.506 & 0.777 & 0.425 & 0.440 \\

        HTS
        & \underline{1796} & 62 & 58.3 & {0.711$_{\pm.023}$}
        & \underline{0.988} & 0.798 & 0.594 & 0.792 & 0.430 & 0.661 \\

        Best-of-$N$ / LTS
        & 5184 & \underline{32} & 145.7 & {0.787$_{\pm.017}$}
        & \underline{0.988} & 0.806 & \underline{0.713} & \underline{0.906} & {0.630} & 0.680 \\

        {FKD~\cite{zhang2026classic}$^{\dagger}$}
        & 5280 & 160 & 154.9 & \underline{0.810}$_{\pm.018}$
        & {\bf 1.000} & \underline{0.859} & {0.707} & {0.894} & \underline{0.690} & \underline{0.710} \\

        \midrule

        \textbf{PATTS (ours)}
        & \underline{1796} & 38.2 & \underline{55.1} & {\bf 0.844}$_{\pm.014}$
        & \textbf{1.000} & \textbf{0.899} & \textbf{0.753} & \textbf{0.915} & \textbf{0.718} & \textbf{0.776} \\

        \bottomrule

    \end{tabular}
\end{table*}
\begin{table}[ht]
    \centering
    \setlength{\tabcolsep}{3pt}
    \renewcommand{\arraystretch}{1.08}
    \caption{
    \textbf{Component breakdown of PATTS on GenEval with MMaDA-Parallel.} The rows progressively add joint thought-image verification, reward routing, and trust estimation. \textbf{Bold} marks the complete method. $N=32$ on full GenEval benchmark.
    }
    \label{tab:patts_component_breakdown}
    \resizebox{\linewidth}{!}{
    \begin{tabular}{lllll|c}
    
    \toprule
    \textbf{Configuration} & \textbf{TTS} & \textbf{Trust Gate} &
    \textbf{Verifier} & \textbf{Reward} &  \textbf{GenEval}$\uparrow$ \\
    \midrule
    
    No TTS &
    \ding{55} & -- & -- & -- & 0.628 \\
    
    + HTS &
    \ding{51} & -- & txt + img & external & 0.711 \\

    $\rightarrow$ Trust-Gated Beam &
    \ding{51} & fixed phase & txt + img & external & 0.773 \\
        
    + hard two-phase reward &
    \ding{51} & fixed phase & txt + img & hard switch & 0.795 \\
        
    + timestep trust gate $c_t:=t$ &
    \ding{51} & timestep & txt + img & mix & {0.812} \\
        
    {\bf $\rightarrow$ full trust gate (PATTS) } &
    \ding{51} & $c_t(h_t)$ & txt + img & mix & {0.844} \\

    \bottomrule
    \end{tabular}
    }
\end{table}

\paragraph{TTS search methods.}
We compare final-image selection (best-of-$N$ over completed thought-image pairs) with trajectory-level search that scores the predicted clean state $\hat{z}_0^{(t)}$ at intermediate cut points. PATTS is the trajectory-level variant whose reliability-calibrated trust gate routes verification weight between internal trajectory signals and the external VLM verifier. Unless noted, trajectory-level runs use $N=32$ particles.

\subsection{Trajectory-gated TTS on MMaDA-Parallel}
\label{sec:exp_main}

\paragraph{Overall performance and efficiency.} Table~\ref{tab:main_tts_comparison} compares PATTS with representative test-time scaling baselines under MMaDA-Parallel. PATTS achieves the best overall GenEval score of 0.844, improving over No TTS (0.628), HTS (0.711), Best-of-$N$/LTS (0.787), and FKD (0.810). Relative to the strongest baseline, FKD, PATTS improves or keeps accuracy across all subtasks. PATTS is also substantially more compute-efficient, with 1796 generator NFEs, compared with 5184 for Best-of-$N$ and 5280 for FKD, while reducing end-to-end latency from 154.9s for FKD to 55.1s. These results indicate that the gain does not arise simply from generating or verifying more candidates, but from allocating test-time computation more selectively along the joint thought-image trajectory.

\paragraph{Performance gain from PATTS components.} 
Table~\ref{tab:patts_component_breakdown} provides a controlled ablation of the complete PATTS result on the full GenEval benchmark with $N=32$. Starting from the no-TTS score of $0.628$, applying trajectory-level HTS with joint text-image verification increases performance to $0.711$. Introducing trust-gated beam further raises the score to $0.773$. Replacing direct external verification with a hard two-phase internal/external reward switch gives $0.795$. A continuous reward mixture controlled only by the denoising timestep, $c_t:=t$, reaches $0.812$. Finally, replacing this fixed timestep schedule with the proposed state-dependent trust $c_t(h_t)$ increases performance to $0.844$. Trajectory-level search, reward routing, and state-dependent trust estimation contribute successively to the final result of overall $0.628\rightarrow0.844$ improvement. In particular, holding the verifier input and continuous reward mixture fixed, replacing timestep-only trust with $c_t(h_t)$ gives a $3.2$ percentage-point improvement, directly isolating the contribution of the proposed trust gate.


\newcommand{\GENoTTSMean}{0.628}
\newcommand{\GENoTTSStd}{0.037}

\newcommand{\GEHTSMean}{0.714}
\newcommand{\GEHTSStd}{0.023}

\newcommand{\GEBestNMean}{0.787}
\newcommand{\GEBestNStd}{0.017}

\newcommand{\GEFKDMean}{0.810}
\newcommand{\GEFKDStd}{0.018}

\newcommand{\GEPATTSMean}{0.844}
\newcommand{\GEPATTSStd}{0.014}

\newcommand{\TCNoTTSMean}{0.4099}
\newcommand{\TCNoTTSStd}{0.02}

\newcommand{\TCHTSMean}{0.4459}
\newcommand{\TCHTSStd}{0.03}

\newcommand{\TCBestNMean}{0.4670}
\newcommand{\TCBestNStd}{0.03}

\newcommand{\TCFKDMean}{0.4914}
\newcommand{\TCFKDStd}{0.015}

\newcommand{\TCPATTSMean}{0.5157}
\newcommand{\TCPATTSStd}{0.01}

\newcommand{\DPGNoTTSMean}{0.7312}
\newcommand{\DPGNoTTSStd}{0.03}

\newcommand{\DPGHTSMean}{0.7505}
\newcommand{\DPGHTSStd}{0.02}

\newcommand{\DPGBestNMean}{0.7826}
\newcommand{\DPGBestNStd}{0.03}

\newcommand{\DPGFKDMean}{0.7998}
\newcommand{\DPGFKDStd}{0.01}

\newcommand{\DPGPATTSMean}{0.8291} 
\newcommand{\DPGPATTSStd}{0.02}     

\definecolor{methodnotts}{HTML}{9DA5AE}
\definecolor{methodfkd}{HTML}{f0ad4b}
\definecolor{methodhts}{HTML}{f25142}
\definecolor{methodbestn}{HTML}{7185f9}
\definecolor{methodpatts}{HTML}{84b171}

\newcommand{\methodbar}[4]{%
    \addplot+[
        bar shift=0pt,
        fill=#4,
        fill opacity=0.8,
        color=#4,
        draw opacity=0.9,
        forget plot,
        error bars/.cd,
        y dir=both,
        y explicit,
        error bar style={color=#4!75!black, opacity=1.0}
    ] coordinates {(#1,#2) +- (0,#3)};%
}

\begin{figure}[ht]
    \centering
    \begin{subfigure}[t]{0.32\linewidth}
        \centering
        \begin{tikzpicture}
        \begin{axis}[
        ybar,
        scale only axis,
        width=0.82\linewidth,
        height=3.2cm,
        bar width=8pt,
        symbolic x coords={No TTS, HTS, Best-$N$, FKD, PATTS},
        xtick=\empty,
        ylabel style={font=\small},
        tick label style={font=\scriptsize},
        ymin=0, ymax=1.0,
        enlarge x limits=0.12,
        nodes near coords,
        nodes near coords style= {font=\tiny, text=black, yshift=2pt,
            /pgf/number format/fixed, /pgf/number format/precision=2},
        legend to name=benchmarkmethodlegend,
        legend columns=5,
        legend style={draw=none, fill=none, font=\scriptsize,
            /tikz/every even column/.append style={column sep=9pt}},
        ]
        
        \addlegendimage{area legend, fill=methodnotts, fill opacity=0.8, draw=methodnotts}
        \addlegendentry{No TTS}

        \addlegendimage{area legend, fill=methodhts, fill opacity=0.8, draw=methodhts}
        \addlegendentry{HTS}
        
        \addlegendimage{area legend, fill=methodbestn, fill opacity=0.8, draw=methodbestn}
        \addlegendentry{Best-of-$N$}
        
        \addlegendimage{area legend, fill=methodfkd, fill opacity=0.8, draw=methodfkd}
        \addlegendentry{FKD}
        
        \addlegendimage{area legend, fill=methodpatts, fill opacity=0.8, draw=methodpatts}
        \addlegendentry{PATTS}

        \methodbar{No TTS}{\GENoTTSMean}{\GENoTTSStd}{methodnotts}
        \methodbar{FKD}{\GEFKDMean}{\GEFKDStd}{methodfkd}
        \methodbar{HTS}{\GEHTSMean}{\GEHTSStd}{methodhts}
        \methodbar{Best-$N$}{\GEBestNMean}{\GEBestNStd}{methodbestn}
        \methodbar{PATTS}{\GEPATTSMean}{\GEPATTSStd}{methodpatts}
        
        \end{axis}
        \end{tikzpicture}
        
        \caption{GenEval}
        \label{fig:benchmark_geneval}
        
    \end{subfigure}
    \hfill
    \begin{subfigure}[t]{0.32\linewidth}
        \centering
        \begin{tikzpicture}
        \begin{axis}[
            ybar,
            scale only axis,
            width=0.82\linewidth,
            height=3.2cm,
            bar width=8pt,
            symbolic x coords={No TTS, HTS, Best-$N$, FKD, PATTS},
            xtick=\empty,
            ylabel style={font=\small},
            tick label style={font=\scriptsize},
            ymin=0, ymax=0.75,
            enlarge x limits=0.12,
            nodes near coords,
            nodes near coords style={font=\tiny, text=black, yshift=2pt,
            /pgf/number format/fixed,
            /pgf/number format/precision=2},
        ]
        \methodbar{No TTS}{\TCNoTTSMean}{\TCNoTTSStd}{methodnotts}
        \methodbar{FKD}{\TCFKDMean}{\TCFKDStd}{methodfkd}
        \methodbar{HTS}{\TCHTSMean}{\TCHTSStd}{methodhts}
        \methodbar{Best-$N$}{\TCBestNMean}{\TCBestNStd}{methodbestn}
        \methodbar{PATTS}{\TCPATTSMean}{\TCPATTSStd}{methodpatts}
        \end{axis}
        \end{tikzpicture}
        \caption{T2I-CompBench}
        \label{fig:benchmark_t2icomp}
    \end{subfigure}
    \hfill
    \begin{subfigure}[t]{0.32\linewidth}
        \centering
        \begin{tikzpicture}
        \begin{axis}[
        ybar,
        scale only axis,
        width=0.82\linewidth,
        height=3.2cm,
        bar width=8pt,
        symbolic x coords={No TTS, HTS, Best-$N$, FKD, PATTS},
        xtick=\empty,
        ylabel style={font=\small},
        tick label style={font=\scriptsize},
        ymin=0, ymax=1.0,
        enlarge x limits=0.12,
        nodes near coords,
        nodes near coords style={font=\tiny, text=black, yshift=2pt,
            /pgf/number format/fixed, /pgf/number format/precision=2},
        ]
        \methodbar{No TTS}{\DPGNoTTSMean}{\DPGNoTTSStd}{methodnotts}
        \methodbar{FKD}{\DPGFKDMean}{\DPGFKDStd}{methodfkd}
        \methodbar{HTS}{\DPGHTSMean}{\DPGHTSStd}{methodhts}
        \methodbar{Best-$N$}{\DPGBestNMean}{\DPGBestNStd}{methodbestn}
        \methodbar{PATTS}{\DPGPATTSMean}{\DPGPATTSStd}{methodpatts}
        \end{axis}
        \end{tikzpicture}
        \caption{DPG-Bench}
        \label{fig:benchmark_dpg}
    \end{subfigure}
    \pgfplotslegendfromname{benchmarkmethodlegend}
    \caption{\textbf{Cross-benchmark comparison of test-time scaling methods.} 
    Overall scores on GenEval, T2I-CompBench, and DPG-Bench under the same MMaDA-Parallel backbone; error bars denote standard deviation from four runs.}
    \label{fig:cross_benchmark_comparison}
\end{figure}
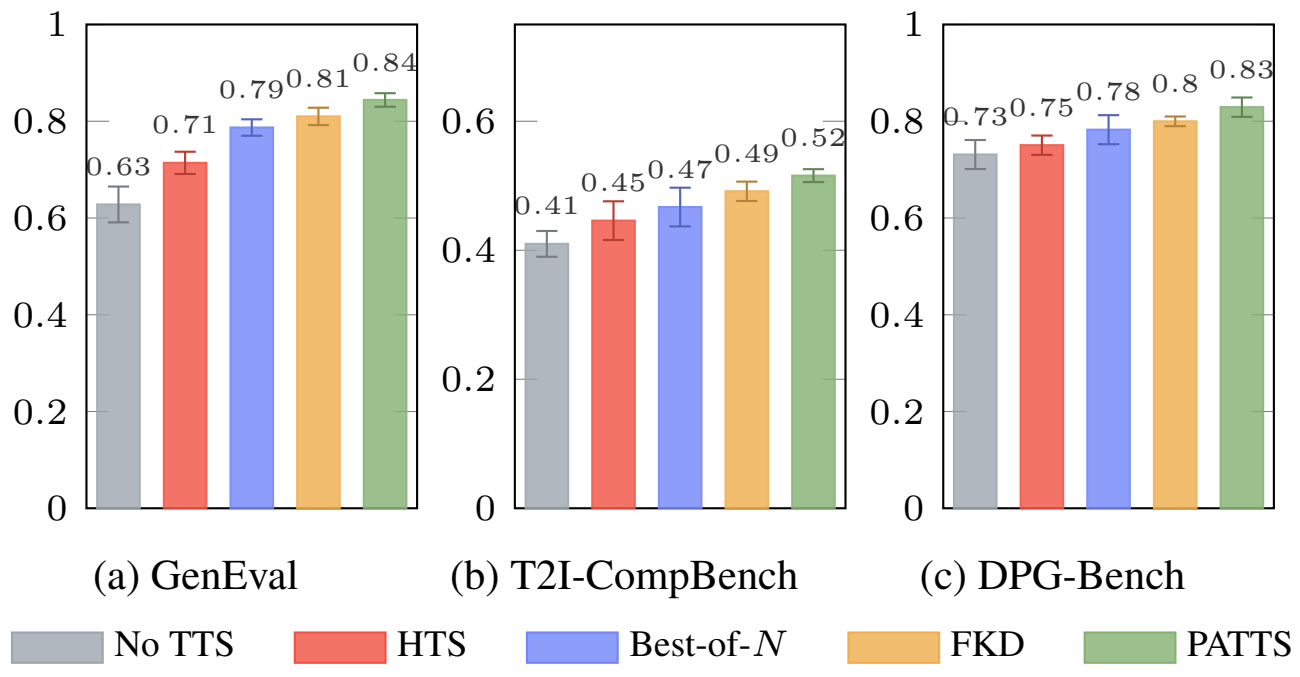

\paragraph{Cross-benchmark robustness.}
Figure~\ref{fig:cross_benchmark_comparison} evaluates whether the advantage persists beyond GenEval on two additional compositional-generation benchmarks. PATTS obtains the highest mean score on GenEval, T2I-CompBench, and DPG-Bench. Together with the error bars, these results provide evidence that reliability-weighted trajectory allocation is not specific to the GenEval evaluator or to a single compositional category. Because all full PATTS experiments use the MMaDA-Parallel backbone, we interpret this as benchmark-level robustness rather than model-level generalization.

\begin{figure}
    \centering
    \includegraphics[width=\linewidth]{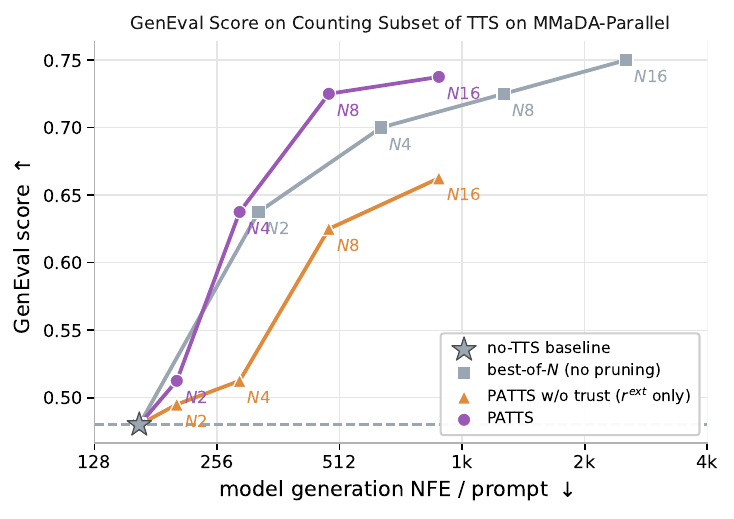}
    \caption{
    \textbf{Test-time compute scaling on the GenEval counting subset.} GenEval score versus model-generation NFEs per prompt for MMaDA-Parallel. PATTS remains competitive with external-only selection across the evaluated budgets, while both curves flatten as the budget grows.
    }
    \label{fig:nfe_scale}
\end{figure}

\paragraph{Scaling with test-time compute.}
Figure~\ref{fig:nfe_scale} further investigates PATTS trajectory-search scaling versus baselines on the GenEval counting subset (we provide further details in the supplementary material). Across the evaluated budgets, PATTS improves as the particle budget increases and remains competitive with or above external-only trajectory selection at matched generation cost. The curves flatten at larger budgets, indicating diminishing returns from additional candidates. This behavior is consistent with the verifier-alignment analysis in Sec.~\ref{sec:exp_failure_modes}: once the candidate pool contains sufficiently strong samples, further scaling becomes increasingly limited by the ability of the verifier to rank them rather than by candidate generation alone.

\subsection{Two failure modes behind the comparison}
\label{sec:exp_failure_modes}

The performance pattern in Sec.~\ref{sec:exp_main} is explained by two properties of final-image \tts that trajectory gating is designed to address.

\begin{figure}[t]
    \centering
    \includegraphics[width=\linewidth]{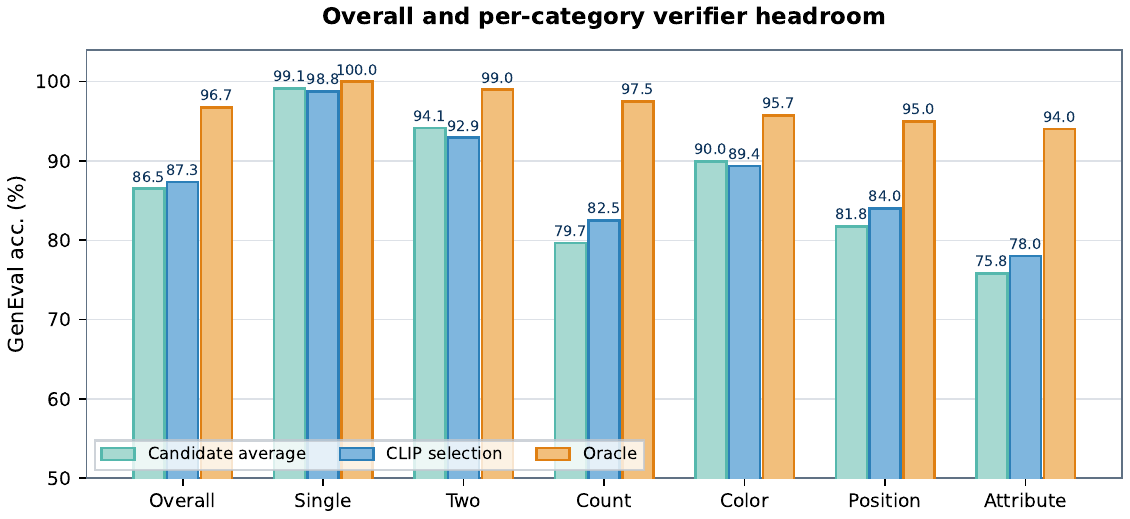}
    \caption{\textbf{Per-category verifier headroom} on Lumina-DiMOO $N=16$ beam-search outputs. Bars show candidate-pool average correctness, CLIP-selected correctness, and the oracle success rate when any candidate for the prompt is correct.}
    \label{fig:headroom}
\end{figure}

\paragraph{Failure mode 1: final-image verifiers are weakly aligned with compositional correctness.}
A final-image verifier can raise selected-image accuracy by preferring visually plausible samples while remaining poorly calibrated to compositional correctness. We test this on Lumina-DiMOO beam-search outputs ($553$ prompts, $N=16$): for each prompt we compare the candidate-pool average, the CLIP-selected image, and an oracle that succeeds if any candidate for the prompt is correct (Fig.~\ref{fig:headroom}). The candidate pool is already strong ($86.51\%$ correct on average); CLIP selection reaches $87.34\%$, only $0.83$ points above the candidate average, while the oracle reaches $96.75\%$. Most of the selected accuracy therefore originates in the candidate pool rather than in verifier ranking, and the $9.41$-point gap to the oracle bounds the headroom available to a better-aligned verifier. This gap is largest for attribute prompts ($16.0$ points).

\begin{table}[t]
    \centering
    \small
    \caption{{\bf MMaDA-Parallel T2I vs.\ thinking-aware generation.}}
    \label{tab:mmada_parallel_thinking}
    \setlength{\tabcolsep}{3pt}
    \renewcommand{\arraystretch}{1.05}
    \resizebox{\linewidth}{!}{
    \begin{tabular}{llllllll}
    \toprule
    \textbf{Setting} & \textbf{Single} & \textbf{Two} & \textbf{Count} & \textbf{Color} & \textbf{Pos.} & \textbf{Attr.} & \textbf{Overall} \\
    \midrule
    Thinking-Aware              & {\bf 0.97} & {\bf 0.72} & {\bf0.48} & {\bf0.70} & 0.31 & 0.54 & 0.62 \\
    Standard T2I (w/o thinking) & 0.95 & 0.71 & 0.43 & {\bf0.70} & {\bf0.57} & {\bf0.56} & {\bf0.65} \\
    \midrule
    $\Delta$ (think $-$ std)  & $0.02$ & $0.01$ & $0.05$ & $0.00$ & $-0.26$ & $-0.02$ & $-0.03$ \\
    \bottomrule
    \end{tabular}
    }
\end{table}

The score-level statistics are consistent with this reading. For ranking candidates within a prompt, the CLIP score has ROC-AUC $0.464$, point-biserial correlation $r{=}-0.046$, and a mean score gap of $-0.005$ between correct and incorrect candidates; on two-object prompts it is anti-correlated with correctness (AUC $0.382$). The weakness is specific to the final-image CLIP score rather than to verification in general: a structured matching score that mirrors GenEval's object-level checks separates correct from incorrect images substantially better.
The same limitation is visible on SD 1.5, detailed in supplementary material.

\paragraph{Failure mode 2: unverified parallel thinking can introduce semantic drift.}
Table~\ref{tab:mmada_parallel_thinking} shows that enabling the reasoning trace without verification slightly reduces overall GenEval accuracy, from 0.65 to 0.62. The degradation is not uniform: single-object, two-object, counting, and color prompts are largely preserved or slightly improved, while position and attribute prompts degrade. This suggests that the reasoning trace does not simply reduce generation quality globally. Instead, it can perturb fine-grained semantic binding and spatial grounding in cases where the text and image states are not well aligned.

\begin{figure}[ht]
    \centering
    \includegraphics[width=\linewidth]{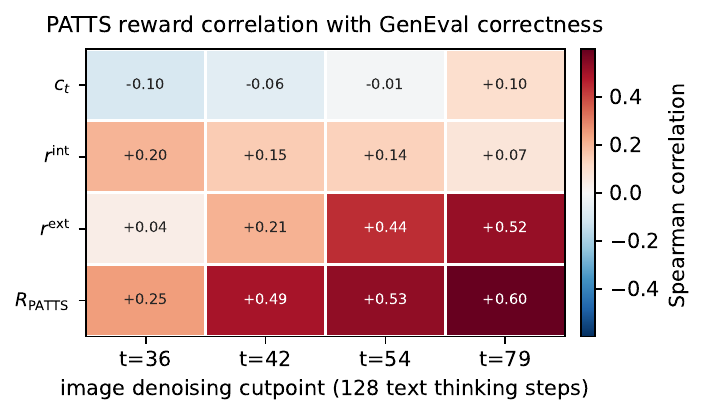}
    \caption{{\bf Correlation Test between GenEval correctness and reward signals.} We Spearman-correlate every selection signal against whether that prompt's \emph{final} image is GenEval-correct ($N=16$, cutpoints shown at text-thinking step $t$ of $128$).}
    \label{fig:reward_corr}
\end{figure}

\begin{table*}[t!]
    \centering
    \footnotesize
    \caption{
    \textbf{Ablation study of PATTS designs.}
    ``{\bf RankAcc}'': intermediate-final ranking agreement, ``{\bf Regret}'':  loss from selecting a suboptimal intermediate state,
    ``{\bf Score/Call}'': verifier-call efficiency. Defaults are shown in the bottom row.
    Evaluated on GenEval counting subset at $N=16$.
    }
    \label{tab:patts_ablation}
    
    \begin{subtable}[t]{0.24\linewidth}
        \centering
        \begin{tabular}{ll}
            \toprule
            {\bf Guidance Signal} & {\bf GenEval$\uparrow$} \\
            \midrule
            no TTS                & 0.50 \\
            external only         & 0.62 \\
            internal only         & 0.45 \\
            hard two-phase        & 0.58 \\
            \arrayrulecolor{black!40}\midrule
            reliability mixture   & {\bf 0.66} \\
            \arrayrulecolor{black}\bottomrule
        \end{tabular}
        \caption{\footnotesize\textbf{Reward routing.} We compare external / internal trajectory verification, hard phase switching, and the proposed mixture.}
        \label{subtab:ablate_guidance}
    \end{subtable}
    \hfill
    \begin{subtable}[t]{0.24\linewidth}
        \centering
        \begin{tabular}{ll}
            \toprule
            {\bf Trust Features} & {\bf RankAcc$\uparrow$} \\
            \midrule
            timestep only         & 0.17 \\
            $H_t,M_t,\rho_t$      & 0.02 \\
            $+S_t$                & 0.07 \\
            $+A_t$                & 0.13 \\
            \arrayrulecolor{black!40}\midrule
            $+{\rm RankStab}_t$   & {\bf 0.19} \\
            \arrayrulecolor{black}\bottomrule
        \end{tabular}
        \caption{\footnotesize\textbf{Trust estimation.} We ablate the state-dependent trust gate $c_t(z_t)$ from timestep-only tempering to full multi-trajectory reliability estimation.}
        \label{subtab:ablate_trust}
    \end{subtable}
    \hfill
    \begin{subtable}[t]{0.24\linewidth}
        \centering
        \begin{tabular}{ll}
            \toprule
            {\bf Selection Score} & {\bf Regret$\downarrow$} \\
            \midrule
            $r_t^{\rm ext}$       & 0.21 \\
            $c_t r_t^{\rm ext}$   & 0.20 \\
            $r_t^{\rm int}$       & 0.34 \\
            $R_t^{\rm PATTS}$     & 0.22 \\
            \arrayrulecolor{black!40}\midrule
            $R_t^{\rm PATTS}-\beta u_t$ & {\bf 0.19} \\
            \arrayrulecolor{black}\bottomrule
        \end{tabular}
        \caption{\footnotesize\textbf{Particle scoring.} We test whether the internal fallback and penalty improve selection decisions. Regret is vs.\ a held-out terminal-verifier oracle.}
        \label{subtab:ablate_selection}
    \end{subtable}
    \hfill
    \begin{subtable}[t]{0.24\linewidth}
        \centering
        \begin{tabular}{ll}
            \toprule
            {\bf Controller} & {\bf Score/Call$\uparrow$} \\
            \midrule
            fixed top-$K$        & 0.021 \\
            fixed $\lambda_t$    & 0.034 \\
            ESS control          & 0.027 \\
            rank-stab.    & 0.031 \\
            \arrayrulecolor{black!40}\midrule
            adaptive  & {\bf 0.036} \\
            \arrayrulecolor{black}\bottomrule
        \end{tabular}
        \caption{\footnotesize\textbf{Search controller.} We isolate each online update ($\lambda_t$, uncertainty $\beta$, ESS resampling) over a fixed top-$K$ baseline with identical reward routing.}
        \label{subtab:ablate_controller}
    \end{subtable}
\end{table*}

\subsection{When is a state reliable enough to verify?}
\label{sec:exp_trust}

PATTS applies the external verifier to the predicted clean state $\hat{z}_0^{(t)}$ only when the trajectory is estimated to be reliable, so whether such a moment is identifiable from cheap online features is an empirical question. 
In Figure~\ref{fig:reward_corr}, under hierarchical beam search, at each cut point we take the top-scoring particle per prompt and Spearman-correlate each selection signal against whether that prompt's final image is GenEval-correct. For external VLM reward, the correlations are relatively weak and unstable at early cut points and become stronger and more consistent as denoising proceeds. 
\emph{This is the regime the trust gate is designed to exploit:} it down-weights external verification while the predicted clean state is uninformative and increases its weight once the state stabilizes. The same observation bounds what the gate can achieve on attribute prompts, the category where the late-denoising signal remains weakest (failure mode~1) and where, correspondingly, trust-gated verification yields the smallest gain in Sec.~\ref{sec:exp_main}.

\subsection{Ablations}
\label{sec:exp_ablation}

Table~\ref{tab:patts_ablation} isolates the PATTS components on the GenEval at $N=16$ on counting subset, where the internal trajectory reward is least predictive of final correctness and therefore provides a challenging test for the method. We report mixed outcomes:
For reward routing, the continuous reliability mixture improves over external-only verification from 0.62 to 0.66, while internal-only routing falls below no-search; on this subset the internal reward is thus useful as a gate input but not as a standalone selector.
For trust estimation, the full state-dependent feature set slightly outperforms a timestep-only schedule (rank accuracy $0.19$ vs.\ $0.17$), while partial feature subsets remain below the timestep baseline.
The clearest gain appears in verifier efficiency: the adaptive controller raises score per verifier call from $0.021$ to $0.036$, substantially reducing verifier cost relative to a fixed beam.
Therefore, on this subset, the main measurable benefit of the reliability-calibrated design is improved verifier-call efficiency rather than a further gain in absolute accuracy, consistent with the weak intermediate-reward correlation reported in Sec.~\ref{sec:exp_trust}.

\paragraph{Relation between trust estimation and ranking accuracy.} The modest ranking advantage of the trust features is consistent with the end-to-end results. In Table~\ref{subtab:ablate_trust}, the full features improve intermediate-final rank agreement from $0.17$ to $0.19$ over timestep-only trust. This measures a local ranking property on the $N=16$ counting subset, whereas Table~\ref{tab:patts_component_breakdown} reports terminal accuracy on the full benchmark after repeated trajectory selection. The result therefore suggests that trajectory diagnostics provide a modest improvement over timestep-only reliability estimation, while the larger end-to-end gain accumulates across trajectory search, reward routing, and trust-aware selection.
\section{Conclusion}
\label{sec:conclusion}

We studied test-time scaling for parallel multimodal diffusion LLMs, where reasoning tokens and image tokens co-evolve within a single generation trajectory. Our analysis shows that final-image selection alone can be limited in this setting: CLIP-style prompt-image scores may favour visually plausible samples without reliably ranking compositional correctness. We introduce PATTS, a state-dependent reliability-weighted trajectory search method that uses model-internal diagnostics to determine when predicted clean thought-image states should receive stronger external VLM verification. On GenEval, the complete PATTS configuration improves over final-output reranking in the fixed-$N=32$ design sweep. The ablations further show that reliable intermediate verification remains the central limitation: internal trajectory signals are not semantic verifiers by themselves, and future progress depends on stronger verifier alignment and more reliable intermediate-state scoring.

\newpage
\section*{Acknowledgments}
\noindent This research is funded in part by an ARC (Australian Research Council) Discovery Grant of DP220100800. Prof.~Hongdong Li holds concurrent appointments as a Full Professor with the ANU and as an Amazon Scholar with Amazon (part time). This paper describes work performed at ANU and is not associated with Amazon.

{
    \small
    \bibliographystyle{ieeenat_fullname}
    \bibliography{main}
}

\clearpage
\appendix
\hypersetup{pageanchor=false}
\setcounter{page}{1}
\setcounter{figure}{0}
\setcounter{table}{0}
\renewcommand{\thefigure}{\Alph{figure}}
\renewcommand{\thetable}{\Alph{table}}
\maketitlesupplementary

\section{PATTS Reliability Computation}
\label{app:reliability}

\paragraph{Cutpoints and clean previews.}
MMaDA-Parallel decodes reasoning tokens over $T{=}128$ steps and updates image tokens on a sparser grid of 32 steps, evenly spaced from text step 32 to 127. A cutpoint is right after the $m$-th image update. We use $m\in\{2,4,8,16\}$ and obtain the final selection at $m{=}32$. At each cutpoint, one forward pass of the frozen generator gives logits for every position. Each still-masked position is filled with its argmax token, which gives the clean preview $\hat z_0^{(t)}=(\hat y_0^{(t)},\hat v_0^{(t)})$. Positions that are already unmasked keep their tokens.

\paragraph{Token statistics ($H_t$, $M_t$, $\rho_t$).}
Let $\mathcal M_t$ be the set of positions that are still masked, pooled over the reasoning span and the image span. Let $L$ be the total number of reasoning and image positions. For $j\in\mathcal M_t$, let $p^j$ be the softmax over that position's vocabulary $\mathcal V_j$: the text vocabulary for reasoning positions and the VQ codebook for image positions. Then
\begin{align*}
  H_t&=\frac{1}{|\mathcal M_t|}\sum_{j\in\mathcal M_t}\frac{-\sum_{v}p^j_v\log p^j_v}{\log|\mathcal V_j|},\\
  M_t&=\frac{1}{|\mathcal M_t|}\sum_{j\in\mathcal M_t}\big(p^j_{(1)}-p^j_{(2)}\big),\\
  \rho_t&=\frac{|\mathcal M_t|}{L},
\end{align*}
where $p_{(1)}, p_{(2)}$ are the two largest probabilities. If no position is masked, $H_t$ and $M_t$ become the average of the text and image means over all positions.

\paragraph{Denoising stability ($S_t$).}
Let $\mathrm{ov}(a,b)=\frac{1}{|a|}\sum_j\mathbb 1[a_j=b_j]$ be the fraction of positions where two token sequences agree. The previous preview is that of the same particle, which survived the previous cut. Then
\begin{equation}
    S_t^{(i)}=\tfrac12\,\mathrm{ov}(\hat y_0^{(t)},\hat y_0^{(t-\Delta)})
    +\tfrac12\,\mathrm{ov}(\hat v_0^{(t)},\hat v_0^{(t-\Delta)}),
    \label{eq:stability}
\end{equation}
Here $t-\Delta$ denotes the \emph{previous cutpoint}, and $S_t^{(i)}=0$ at the first cutpoint.

\paragraph{Rank stability ($\mathrm{RankStab}_t$).}
Given each trajectory with a provisional score $q_t^{(i)}=M_t^{(i)}-H_t^{(i)}+S_t^{(i)}$. Let $B$ be the current population size. $\operatorname{rank}_t^{(i)}\in\{0,\dots,B-1\}$ is particle $i$'s rank by $q_t$ among the $B$ particles, and $\operatorname{rank}_{t-\Delta}^{(i)}$ is its rank by the scores $q_{t-\Delta}$ that the same $B$ particles had at the previous cutpoint. Because both ranks are taken over the same population, pruning does not count as instability. Then
\begin{equation}
    \mathrm{RankStab}_t^{(i)}=
    \exp\!\left(-\frac{|\operatorname{rank}_t^{(i)}-\operatorname{rank}_{t-\Delta}^{(i)}|}{B}\right),
    \label{eq:rankstab}
\end{equation}
and $\mathrm{RankStab}_t^{(i)}=0$ at the first cutpoint.

The external reward combines three Qwen2.5-VL-7B answers on image-thjought questions, each scored as $\sigma(\ell_{\rm Yes}-\ell_{\rm No})$. A particle is verified only when $c_t^{(i)}\ge\tau_{\rm ext}$. Otherwise it keeps the score from its last verification. Particles that have never been verified get $\tilde r^{\rm ext}=0$, the population mean. Both rewards are standardized over the current population,
\begin{equation}
  \tilde r^{(i)}=\frac{r^{(i)}-\mu_t}{\sigma_t+\epsilon},\qquad \epsilon=10^{-6},
  \label{eq:standardize}
\end{equation}
where $(\mu_t,\sigma_t)$ for $r^{\rm ext}$ are computed over verified particles only. We keep the top-$K_t$ particles by this score, with $K_t=16,8,4,2$ at the four cutpoints.

\section{Additional Experiment Results}

\paragraph{Example of parallel thought-image generation.}
Figure~\ref{fig:appendix_visual_tts} shows one generated complete reasoning trace together with the image tokens committed at different denoising intervals. The example illustrates the setting studied in the main paper: reasoning and image content are produced jointly, so the intermediate trajectory contains cross-modal information before the final image is decoded.

\paragraph{Baseline Compositional Accuracy.}
Table~\ref{tab:geneval_baselines} reports the baseline GenEval evaluations used to contextualize the main experiments. Lumina-DiMOO is the strongest generator in this group, reaching an overall score of $0.7824$ and improving substantially over SD~1.5 on relation-heavy categories such as position and color attributes. MMaDA is competitive on single-object and color prompts but remains weak on position and attribute binding. Muddit is substantially below the other models in this run; the generation logs indicate that this configuration used too few denoising steps, so we treat it as an under-tuned baseline rather than a definitive estimate of the model family.

\begin{table}[ht]
    \centering
    \resizebox{\linewidth}{!}{
    \small
    \begin{tabular}{lrrrrrrr}
        \toprule
        Model & Single & Two & Count & Color & Pos. & Attr. & Overall \\
        \midrule
        SD~1.5~\cite{rombach2022ldm}      & 95.94 & 39.39 & 36.88 & 74.20 & 3.25 & 4.29 & 0.4232 \\
        Lumina-DiMOO~\cite{xin2025lumina} & 98.75 & 82.58 & 63.44 & 86.44 & 74.00 & 64.25 & 0.7824 \\
        MMaDA~\cite{yang2026mmada}        & 91.87 & 51.01 & 29.38 & 70.48 & 14.25 & 11.25 & 0.4471 \\
        \bottomrule
    \end{tabular}
    }
    \caption{{\bf Baseline GenEval accuracy.} Category columns report accuracy in percent; overall is the average over categories.}
    \label{tab:geneval_baselines}
\end{table}

\paragraph{Selection-based \tts works on conventional generators, but is verifier-limited.}
Table~\ref{tab:sd15_tts} shows that independent sampling followed by verifier selection is a strong baseline for a conventional generator. Best-of-$N$ improves SD~1.5 from $0.4239$ to $0.5392$ at $N=16$ and $0.5444$ at $N=32$. The gains are concentrated in two-object, counting, color, position, and color-attribute prompts, while single-object prompts are already near saturation. The hierarchical HTS variant improves less, reaching $0.4566$ at $N{=}32,K{=}8$; in this setting, allocating compute to a larger set of independent samples is more effective than the beam-style allocation. Consistent with failure mode~1 in the main paper, these CLIP-guided runs improve the selected samples but still leave substantial compositional headroom relative to an oracle selector.

\begin{table}[h]
    \centering
    \resizebox{\linewidth}{!}{
    \small
    \begin{tabular}{llllllllll}
        \toprule
        Method & $N$ & $K$ & Single & Two & Counting & Colors & Position & Attribute & Overall \\
        \midrule
        Baseline & 1 & -- & 97.50 & 37.37 & 35.00 & 74.47 & 7.00 & 3.00 & 0.4239 \\
        LTS & 16 & -- & 98.75 & 71.72 & 50.00 & 84.04 & 10.00 & 9.00 & 0.5392 \\
        LTS & 32 & -- & 100.00 & 72.73 & 50.00 & 81.91 & 11.00 & 11.00 & 0.5444 \\
        HTS & 16 & 4 & 92.50 & 57.58 & 31.25 & 63.83 & 9.00 & 4.00 & 0.4303 \\
        HTS & 32 & 8 & 91.25 & 56.57 & 46.25 & 64.89 & 8.00 & 7.00 & 0.4566 \\
        \bottomrule
    \end{tabular}
    }
    \caption{{\bf SD~1.5 test-time scaling on GenEval.} All methods return one selected image for each of the 553 prompts. Category columns report accuracy in percent.}
    \label{tab:sd15_tts}
\end{table}

\paragraph{Unverified thinking degrades compositional accuracy.}
Table~\ref{tab:mmada_parallel_thinking} in the main paper isolates the effect of enabling the reasoning trace under the active MMaDA-Parallel evaluation used in the main paper. Standard T2I decoding without the reasoning trace reaches a GenEval overall score of $0.65$, while thinking-aware decoding without verification reaches $0.62$ ($-0.03$). The change is category-dependent: single-object, two-object, and counting accuracy are preserved or slightly improved ($0.95\to0.97$, $0.71\to0.72$, and $0.43\to0.48$), color is unchanged, but position accuracy drops sharply ($0.57\to0.31$) and attribute accuracy decreases slightly ($0.56\to0.54$). This supports the main-paper interpretation that unverified parallel thinking does not simply reduce image quality globally; rather, it can introduce semantic drift in spatial grounding and fine-grained binding when the text and image states are not kept aligned.

\paragraph{Scaling with candidate budget.}
For the test-time compute scaling experiment in Figure \ref{fig:nfe_scale}, all methods are evaluated on the GenEval counting subset with MMaDA-Parallel using 128 text-diffusion steps and 32 image diffusion steps. \(N\) denotes the candidate or particle budget, and NFE denotes the mean number of image-generator forward passes per prompt. Best-of-\(N\) independently generates and verifies \(N\) completed candidates. PATTS w/o Trust Gate performs intermediate trajectory selection using external verification only, while PATTS uses the full reliability-gated search. The no-TTS baseline uses 161 NFEs, requiring no verifier calls.

\section{Additional Visual Examples}

\begin{figure*}[t]
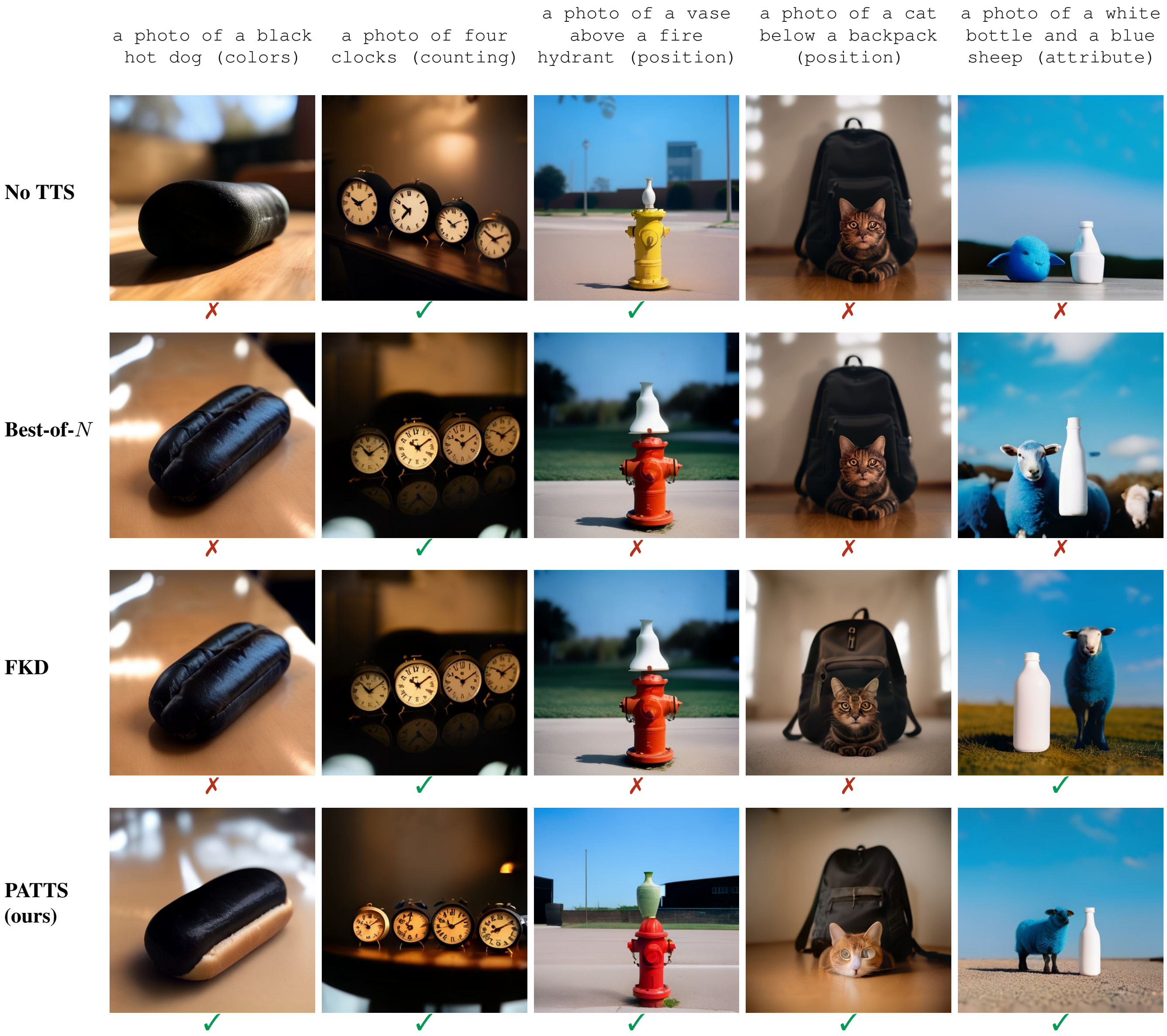

  \centering
  \setlength{\tabcolsep}{1.5pt}
  \renewcommand{\arraystretch}{1.0}
  \begin{tabular}{l ccccc}
    & \visprompt{a photo of a black hot dog (colors)}{}
    & \visprompt{a photo of four clocks (counting)}{}
    & \visprompt{a photo of a vase above a fire hydrant (position)}{}
    & \visprompt{a photo of a cat below a backpack (position)}{}
    & \visprompt{a photo of a white bottle and a blue sheep (attribute)}{} \\ 
    \visrow{No TTS}{no_tts}{\visno}{\visok}{\visok}{\visno}{\visno}
    \visrow{Best-of-$N$}{best_of_n}{\visno}{\visok}{\visno}{\visno}{\visno}
    \visrow{FKD}{fkd}{\visno}{\visok}{\visno}{\visno}{\visok}
    \visrow{PATTS (ours)}{patts}{\visok}{\visok}{\visok}{\visok}{\visok}
  \end{tabular}
  \caption{\textbf{Qualitative comparison with TTS baselines.}
  Final images selected by each method for GenEval prompts with MMaDA-Parallel ($N{=}32$, 128 text / 32 image denoising steps, CFG 4.0). All methods select from the same 32 recorded thought-image trajectories per prompt (FKD's resampling is replayed on this pool); No TTS shows a random trajectory. \visok/\visno\ is the GenEval detector's prediction.}
  \label{fig:appendix_visual_tts}
\end{figure*}

Figure~\ref{fig:app_viusal_parallel} illustrates the full reasoning trace in MMaDA-Parallel. In Figure~\ref{fig:appendix_visual_tts}, Best-of-$N$ and FKD rank candidates by their final or intermediate verifier scores and can prefer images that look plausible but miss the compositional constraint: a black object without the bun, a white cap fused into the hydrant instead of a separate vase, or an extra white sheep. PATTS weights the verifier by the trust gate and selects trajectories whose previews stay stable and cross-modally consistent, which here preserves the counted, placed, and attribute-bound objects.

\definecolor{StepBlue}{RGB}{0,114,178}
\definecolor{StepOrange}{RGB}{230,159,0}
\definecolor{StepGreen}{RGB}{0,158,115}
\definecolor{StepMagenta}{RGB}{204,121,167}

\newcommand{\blue}[1]{\textcolor{StepBlue}{#1 }}
\newcommand{\orange}[1]{\textcolor{StepOrange}{#1 }}
\newcommand{\green}[1]{\textcolor{StepGreen}{#1 }}
\newcommand{\magenta}[1]{\textcolor{StepMagenta}{#1 }}


\begin{figure}[t]
    \centering
    \begin{minipage}{\linewidth}
        \centering
        \includegraphics[width=0.4\linewidth]{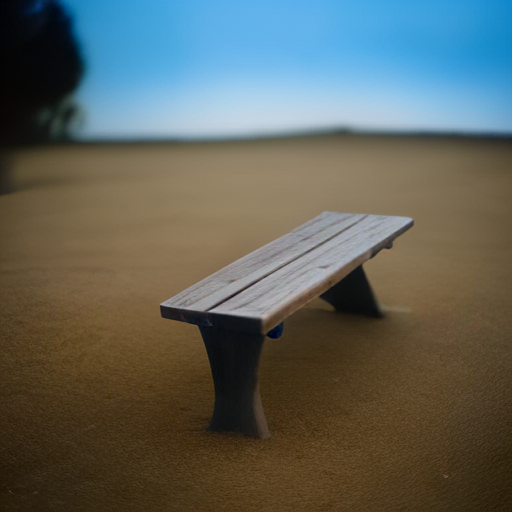}
        \includegraphics[width=0.4\linewidth]{asset/color_masked.png}
    \end{minipage}
    \hfill
    \begin{minipage}{\linewidth}
        \centering
        \captionof{figure}{
        \textbf{Prompt:} \texttt{a photo of a bench}
        \textbf{Color map:}\\
        \blue{\rule{1.2em}{0.8em} steps 0-3}\quad
        \orange{\rule{1.2em}{0.8em} steps 4-7}\quad
        \green{\rule{1.2em}{0.8em} steps 8-11}\quad
        \magenta{\rule{1.2em}{0.8em} steps 12-15}
        \\ 
        \blue{To transform the question image into the answer image, I need to focus on the following aspects:}
        \blue{1. \textbf{Lighting}: The current image has shadows and appears somewhat dimmer compared to the answer image. The answer image has a brighter, more even lighting setup, suggesting an overcast or early morning}
        \orange{ light, which can enhance the mood and}%
        \green{ create}%
        \orange{ a}%
        \green{ more serene atmosphere}%
        \orange{.}
        \orange{2. \textbf{Color}}%
        \green{\textbf{ Tone}}%
        \orange{\textbf{:} The colors in the question image are muted and somewhat}%
        \green{ desaturated}%
        \orange{. The answer image}%
        \green{ has a vibrant blue tint, which}%
        \magenta{ could be due to the sunrisecast sunrise. This will adds a sense of warmth and and a feeling of tranquility to the scene.}
        \magenta{3. \textbf{Background}: The background in the question image is blurred
        and slightly out of focus}%
        \green{. The answer image}%
        \magenta{ has}%
        \green{ a shallow}%
        \magenta{ depth}%
        \green{ of}%
        \magenta{ field, focusing on}%
        \green{ the bench andurring out the field. This can create a sense of depth and
        draw attention to the contrast between the bench and the background}%
        \blue{.}
        \blue{4. }%
        \orange{\textbf{and Texture}}%
        \blue{\textbf{:} The}%
        \orange{ wood}%
        \blue{ in}%
        \orange{ the}%
        \blue{ answer image}%
        \orange{ looks slightly more worn and aged compared to the clean, new bench in
        the question image}%
        \blue{. This can imply a passage over time, enhancing the realism of the scene.}
        \blue{By applying these adjustments, the image image will better reflect the
        transformation.}
        }
        \label{fig:app_viusal_parallel}
    \end{minipage}
\end{figure}

\end{document}